\documentclass[sigconf]{acmart}
\usepackage{graphicx}
\usepackage{amsfonts}
\usepackage{subcaption}
\usepackage{textcomp}
\usepackage{hyperref}
\usepackage{multirow}
\usepackage{orcidlink}

\makeatletter
\def\@ACM@checkaffil{% Only warnings
    \if@ACM@instpresent\else
    \ClassWarningNoLine{\@classname}{No institution present for an affiliation}%
    \fi
    \if@ACM@citypresent\else
    \ClassWarningNoLine{\@classname}{No city present for an affiliation}%
    \fi
    \if@ACM@countrypresent\else
        \ClassWarningNoLine{\@classname}{No country present for an affiliation}%
    \fi
}
\makeatother

\AtBeginDocument{%
  \providecommand\BibTeX{{%
    \normalfont B\kern-0.5em{\scshape i\kern-0.25em b}\kern-0.8em\TeX}}}

\copyrightyear{2023}
\acmYear{2023}
\setcopyright{acmlicensed}\acmConference[SOICT 2023]{The 12th International Symposium on Information and Communication Technology}{December 7--8, 2023}{Ho Chi Minh, Vietnam}
\acmBooktitle{The 12th International Symposium on Information and Communication Technology (SOICT 2023), December 7--8, 2023, Ho Chi Minh, Vietnam}
\acmPrice{15.00}
\acmDOI{10.1145/3628797.3628954}
\acmISBN{979-8-4007-0891-6/23/12}

\begin{document}
\title{Multi-Branch Network for Imagery Emotion Prediction}

% \author{Quoc-Bao Ninh\orcidlink{0009-0002-0341-3540}\textsuperscript{\rm 1,2,*},
% Hai-Chan Nguyen\orcidlink{0009-0007-5732-8671}\textsuperscript{\rm 1,2,*}, Triet Huynh\orcidlink{0009-0007-4277-1881}\textsuperscript{\rm 1,2}, Trung-Nghia Le\orcidlink{0000-0002-7363-2610}\textsuperscript{\rm 1,2,**}
% }

% \affiliation{%
%   \institution{\textsuperscript{\rm 1} Faculty of Information Technology, University of Science, Ho Chi Minh City, Vietnam}
%     % \city{Ho Chi Minh City}
%     % \country{Vietnam}
% }

% \affiliation{%
%   \institution{\textsuperscript{\rm 2} Vietnam National University, Ho Chi Minh City, Vietnam}
%     % \city{Ho Chi Minh City}
%     % \country{Vietnam}
% }

\author{Quoc-Bao Ninh}

\authornotemark[1]
\orcid{0009-0002-0341-3540}
\affiliation{%
  \institution{\textsuperscript{\rm 1} Faculty of Information Technology, University of Science, Ho Chi Minh City, Vietnam}
    % \city{Ho Chi Minh City}
    % \country{Vietnam}
}

\affiliation{
  \institution{\textsuperscript{\rm 2} Vietnam National University, Ho Chi Minh City, Vietnam}
    % \city{Ho Chi Minh City}
    % \country{Vietnam}
}
\email{21120004@student.hcmus.edu.vn}

\author{Hai-Chan Nguyen}
\authornotemark[1]
\orcid{0009-0007-5732-8671}
\affiliation{%
  \institution{\textsuperscript{\rm 1} Faculty of Information Technology, University of Science, Ho Chi Minh City, Vietnam}
    % \city{Ho Chi Minh City}
    % \country{Vietnam}
}

\affiliation{
  \institution{\textsuperscript{\rm 2} Vietnam National University, Ho Chi Minh City, Vietnam}
    % \city{Ho Chi Minh City}
    % \country{Vietnam}
}

\email{21120006@student.hcmus.edu.vn}

\author{Triet Huynh}
\authornotemark[1]
\orcid{0009-0007-4277-1881}
\affiliation{%
  \institution{\textsuperscript{\rm 1} Faculty of Information Technology, University of Science, Ho Chi Minh City, Vietnam}
    % \city{Ho Chi Minh City}
    % \country{Vietnam}
}

\affiliation{
  \institution{\textsuperscript{\rm 2} Vietnam National University, Ho Chi Minh City, Vietnam}
    % \city{Ho Chi Minh City}
    % \country{Vietnam}
}
\email{21120577@student.hcmus.edu.vn}

\author{Trung-Nghia Le}
\authornote{}
\authornotemark[1]
\orcid{0000-0002-7363-2610}
\affiliation{%
  \institution{\textsuperscript{\rm 1} Faculty of Information Technology, University of Science, Ho Chi Minh City, Vietnam}
    % \city{Ho Chi Minh City}
    % \country{Vietnam}
}
\affiliation{
  \institution{\textsuperscript{\rm 2} Vietnam National University, Ho Chi Minh City, Vietnam}
    % \city{Ho Chi Minh City}
    % \country{Vietnam}
}
\email{ltnghia@fit.hcmus.edu.vn}

\thanks{*Equal contributors}
\thanks{**Corresponding author. Email address: ltnghia@fit.hcmus.edu.vn}

\renewcommand{\shortauthors}{Q.-B. Ninh et al.}

\begin{abstract}
For a long time, images have proved perfect at both storing and conveying rich semantics, especially human emotions. A lot of research has been conducted to provide machines with the ability to recognize emotions in photos of people. Previous methods mostly focus on facial expressions but fail to consider the scene context, meanwhile scene context plays an important role in predicting emotions, leading to more accurate results. In addition, Valence-Arousal-Dominance (VAD) values offer a more precise quantitative understanding of continuous emotions, yet there has been less emphasis on predicting them compared to discrete emotional categories. In this paper, we present a novel Multi-Branch Network (MBN), which utilizes various source information, including faces, bodies, and scene contexts to predict both discrete and continuous emotions in an image. Experimental results on EMOTIC dataset, which contains large-scale images of people in unconstrained situations labeled with 26 discrete categories of emotions and VAD values, show that our proposed method significantly outperforms state-of-the-art methods with 28.4\% in mAP and 0.93 in MAE. The results highlight the importance of utilizing multiple contextual information in emotion prediction and illustrate the potential of our proposed method in a wide range of applications, such as effective computing, human-computer interaction, and social robotics. Source code: \url{https://github.com/BaoNinh2808/Multi-Branch-Network-for-Imagery-Emotion-Prediction}
% Our method is trained and evaluated on the EMOTIC dataset, which contains large-scale images of people in unconstrained situations labeled with 26 discrete categories of emotions and VAD values. The proposed method achieves remarkable performance on the EMOTIC dataset with 28.4\% in mAP and 0.93 in MAE. Our method also significantly improves the mAP value by 2.3\% in discrete emotion prediction and reduces the 1.2\% error in VAD values compared to using only contextual and body information. 

\end{abstract}



\maketitle

\section{Introduction}
\label{sec:Intro}
An estimated 4.59 billion people in 2022 are using social media worldwide~\cite{No.User}, with approximately 3.2 billion photos being shared across all social media platforms every day~\cite{No.Image}. Algorithms analyze and process large amounts of information from images, enabling social media platforms to develop strategies that effectively attract users for increasing profits~\cite{Social_Algorithms_1}. Among them, analyzing the emotions and moods of humans in images plays an extremely important role~\cite{haleem2022artificial}. It helps improve the user experience, analyze user behavior and trends, and develop appropriate marketing strategies and advertising campaigns. Therefore, it poses both motivation and challenges for the field of emotion analysis from images, particularly human images.

\begin{figure}[t!]
\centering
\begin{subfigure}[b]{0.4\columnwidth}
  \includegraphics[height=2.3cm, width = \columnwidth]{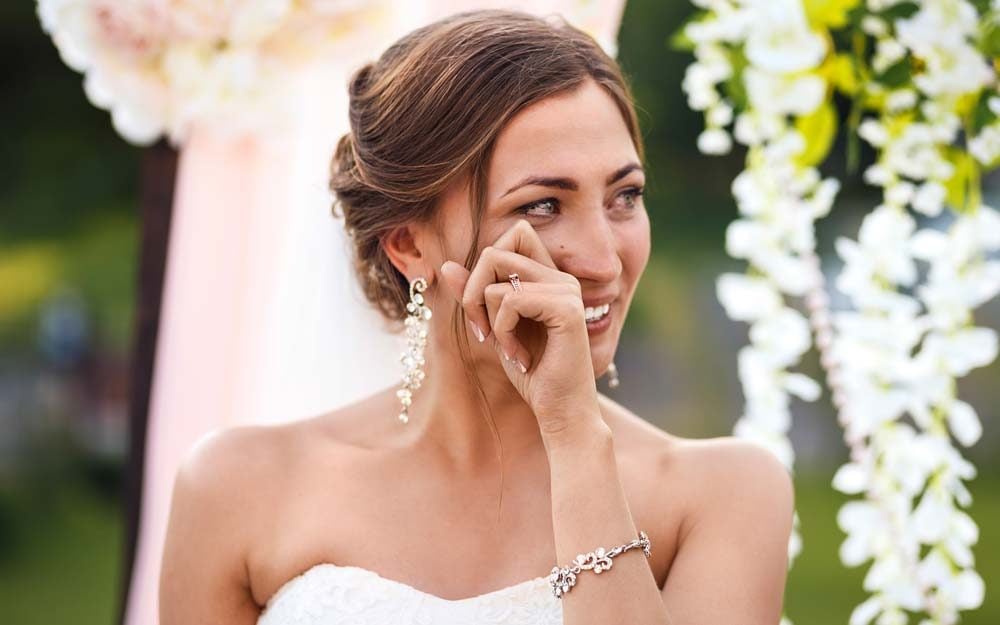}
  \caption{Crying due to joy on her wedding day.}
  \label{fig:image1}
\end{subfigure}
\hspace{0.5cm}
\begin{subfigure}[b]{0.4\columnwidth}
  \includegraphics[height=2.3cm, width = \columnwidth]{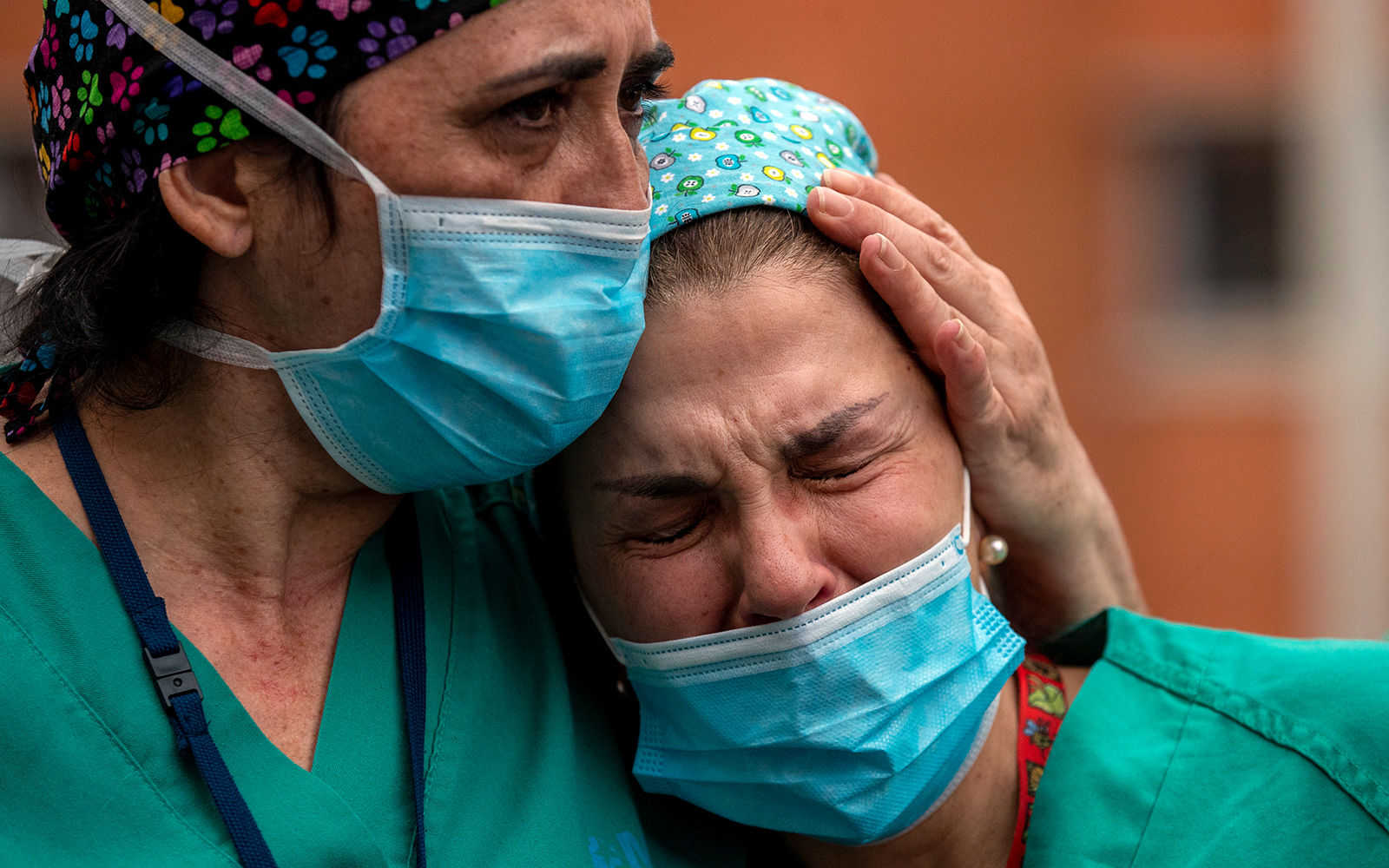}
  \caption{Crying due to the death of a COVID-19 patient.}
  \label{fig:image2}
\end{subfigure}
\caption{Crying in different context.}
\label{fig:comparison}
\end{figure}

Deep learning is a transformative field of machine learning that uses neural networks to solve complex problems. It has revolutionized computer vision, natural language processing, and speech recognition\cite{dong2021survey}. Deep learning models have the ability to efficiently learn features from images, enabling them to achieve high effectiveness in object recognition and classification tasks. Therefore, in the past decade, deep learning has been widely applied in research and development of applications in the field of emotion recognition and analysis from images.
Recently, Transformer models have expanded their applications from natural language processing to computer vision, demonstrating outstanding performance in emotion prediction of image and opening up a new direction for deep learning models in this field.

This paper focuses on the issue of predicting emotions conveyed in images based on contextual information. Concretely, we aim to identify the apparent emotional state of a person depicted in a given image. This problem has been extensively studied in the field of computer vision, with a particular emphasis on two approaches: analyzing facial expressions and analyzing body posture and gestures. These approaches have yielded notable results and practical applications in real-life scenarios~\cite{noroozi2018survey}. However, they only concentrate on analyzing the information expressed by the human in the images, neglecting other equally important information, such as the contextual background of the image. The importance of the contextual background of the image is demonstrated in ~\cite{barrett2011context}, ~\cite{barrett2017emotions}. For example, in Fig. \ref{fig:comparison}, the person may be crying tears of joy at a wedding day. However, on the contrary, the person may also be crying out of frustration or anger due to the devastating impact of the Covid-19 pandemic. Therefore, considering the images shared on social media platforms today, which often encompass both human and their surrounding contexts, solely relying on these traditional approaches to predict emotions may not fulfill the need for high-quality information on social media platforms.

\begin{figure}[t!]
    \centering
    \includegraphics[width=\columnwidth]{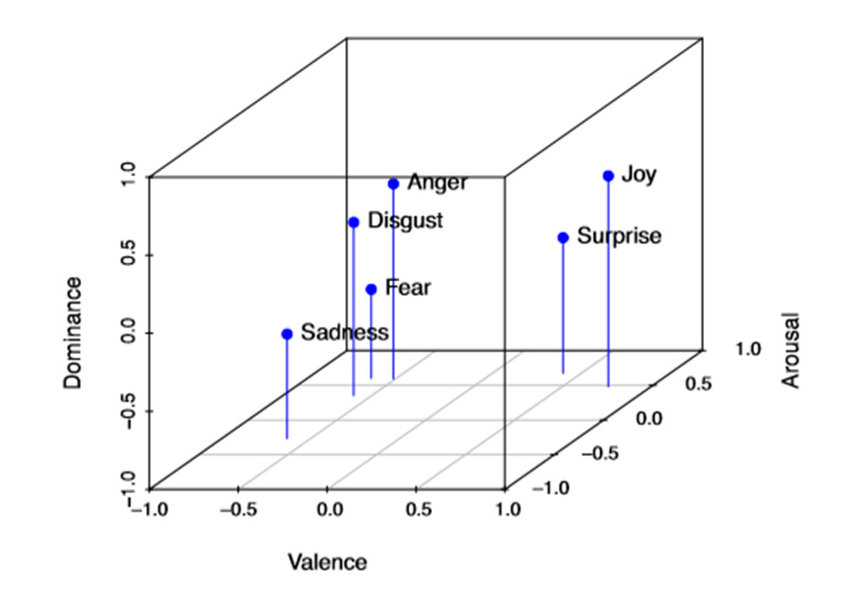}
    \caption{Distribution of the six emotions in valance-arousal-dominance (V.A.D) space ~\cite{VADpic}.}
    \label{fig:EmotionInVAD}
\end{figure}

Emotions can be identified by discrete categories or represented as coordinates on a multi-dimensional continuous scale. Classifying emotions into different categories is convenient for ease of expression and visualization, but it only represents emotions in a qualitative manner and lacks information~\cite{young1997facial}. Even for the same type of emotion, its intensity can vary depending on different situations. Furthermore, an emotion does not solely belong to a single category and can belong to multiple categories, making it difficult and ambiguous to determine. Using a multi-dimensional continuous scale, such as Valence-Arousal-Dominance (VAD) (see Fig.~\ref{fig:EmotionInVAD}), provides more information about a specific emotion, such as its pleasantness, arousal level, and level of control~\cite{harmon2017importance}. For example, an emotion such as anger may have a low value for valence (unpleasant), a high level of arousal (extremely stimulating), and a high level of dominance (a feeling of being hard to control). This approach quantifies emotions and helps to determine them more accurately, making it convenient for computational or further processing tasks. However, it is less commonly used due to its complexity and lack of popularity. In addition, evaluating emotions on a continuous scale is often overlooked due to the lack of labeled data on the continuous scale, whereas there is a wealth of labeled data for discrete emotions in image datasets.

% Most image emotion prediction models are built to recognize discrete emotions, i.e., they perform classification tasks. Nonetheless, evaluating emotions on a continuous scale is often overlooked, even though it can provide more accurate identification of emotions and yield more information for computational and further processing tasks. One of the reasons for this is the lack of labeled data on the continuous scale, whereas there is a wealth of labeled data for discrete emotions in image datasets. With the emergence of labeled datasets on a continuous scale, such as EMOTIC, OASIS, GRAPED, etc., it is necessary to build models that predict the values of Valence-Arousal-Dominance (VAD) for emotions.

To overcome the limitations of traditional methods in predicting human emotions, we propose an effective yet straightforward solution. Our proposed Multi-Branch Network (MBN) consists of three branches that analyze various contextual information, such as person's face, and body, and scene context. These extracted features significantly contribute to the prediction of both discrete and continuous emotions.

Experiments conducted on the EMOTIC dataset~\cite{kosti2017emotic} show that our MBN significantly outperforms the state-of-the-art method~\cite{8713881}. Particularly, the proposed method achieved 0.2837 and 0.9256 in terms of mAP and MAE on the EMOTIC dataset, respectively. We also analyzed the performance of MBN on different backbones.

% the highest mAP result of 0.2837 and the lowest MAE of 0.9256. Beside, the experimental results show that our proposed method improved the accuracy of predicting discrete emotions, measured by mAP (Mean Average Precision), by 2.3\% and reduced the mean absolute error (MEA) by 1.2\% when predicting the VAD values of emotions, compared to the framework presented in .

% inspired by the research presented in article~\cite{8713881} to predict both discrete and continuous emotions. 

Our contributions are as follows:
\begin{itemize}
    \item We present an effective method for human emotion prediction from images. Our method integrates information extracted from various sources, including person's face and body, and scene context. The combination of facial features and attention to scene contexts helps improve the performance of both discrete and continuous emotion prediction. %Our method achieves the state-of-the-art performance on the EMOTIC dataset.

    \item We provide extensive experiments and analysis on emotion prediction using the continuous VAD scale, which can serve as a baseline for future studies. Experimental results contribute to the limited existing research on emotion recognition models using the continuous VAD scale.
\end{itemize}

% The remained structure of this paper is as follows: We mention related works in Section \ref{sec:II}. Section \ref{sec:III} describes our proposed method. Section \ref{sec:IV} discusses experimental results. Finally, in Section \ref{sec:V}, we provide a summary of our conclusions and suggestions for future research.

\section{Related work}
\label{sec:II}
\subsection{Imagery Emotion Prediction}
Emotional prediction has been a long-standing area of research. In 1967, Mehrabian showed that 7\% of knowledge moves between people verbally, 38\% through voice, and 55\% through facial expression~\cite{marechal2019survey}. Facial expressions have long been considered a key component in understanding and predicting human emotions. The most successful CNN models in this field commonly employ popular backbones such as ResNet~\cite{wen2023distract} and EfficientNet~\cite{savchenko2021facial}. In recent years, with the emergence of Transformer models, a new approach has been created with great potential for development in the field of facial expression recognition. Jia Li, et al. achieved high results on the AffectNet dataset by using Vision Transformer~\cite{ma2021facial}.

Another approach to emotion prediction is body gesture and posture;  researchers have also turned to emotional body language, i.e., the expression of emotions through human body pose and body motion ~\cite{schindler2008recognizing}. An implicit assumption common to the work on emotional body language is that body language is only a different means of expressing the same set of basic emotions as facial expressions. Other studies have combined the recognition of body gestures, posture, and facial expressions to achieve more accurate prediction results~\cite{behera2020associating}.

However, we cannot predict exactly the emotions of an image solely based on the facial expressions or body gestures of people, but rather have to take into account the context of the image. A study conducted by Google AI and UC Berkeley analyzed the contexts in which people use a variety of facial expressions across 144 countries~\cite{googleaiblog2021understanding}. They found that each kind of facial expression had distinct associations with a set of contexts that were 70\% preserved across 12 world regions . This suggests that context plays a significant role in how facial expressions are interpreted and understood. Several studies have shown that incorporating contextual features into models can improve their accuracy in predicting emotions. For example, a study on sentiment and emotion classification using bidirectional LSTM found that by incorporating contextual information among utterances in the same video, their proposed attention-based multimodal contextual fusion model outperformed existing methods by over 3\% in emotion classification ~\cite{huddar2021attention}. Another study found that perceives make use of a wide variety of perceptible cues including social context, facial expression, and tone of voice to infer what emotions others are feeling ~\cite{barrett2016emotional}.

\begin{figure*}[t!]
    \centering
    \includegraphics[width=\textwidth]{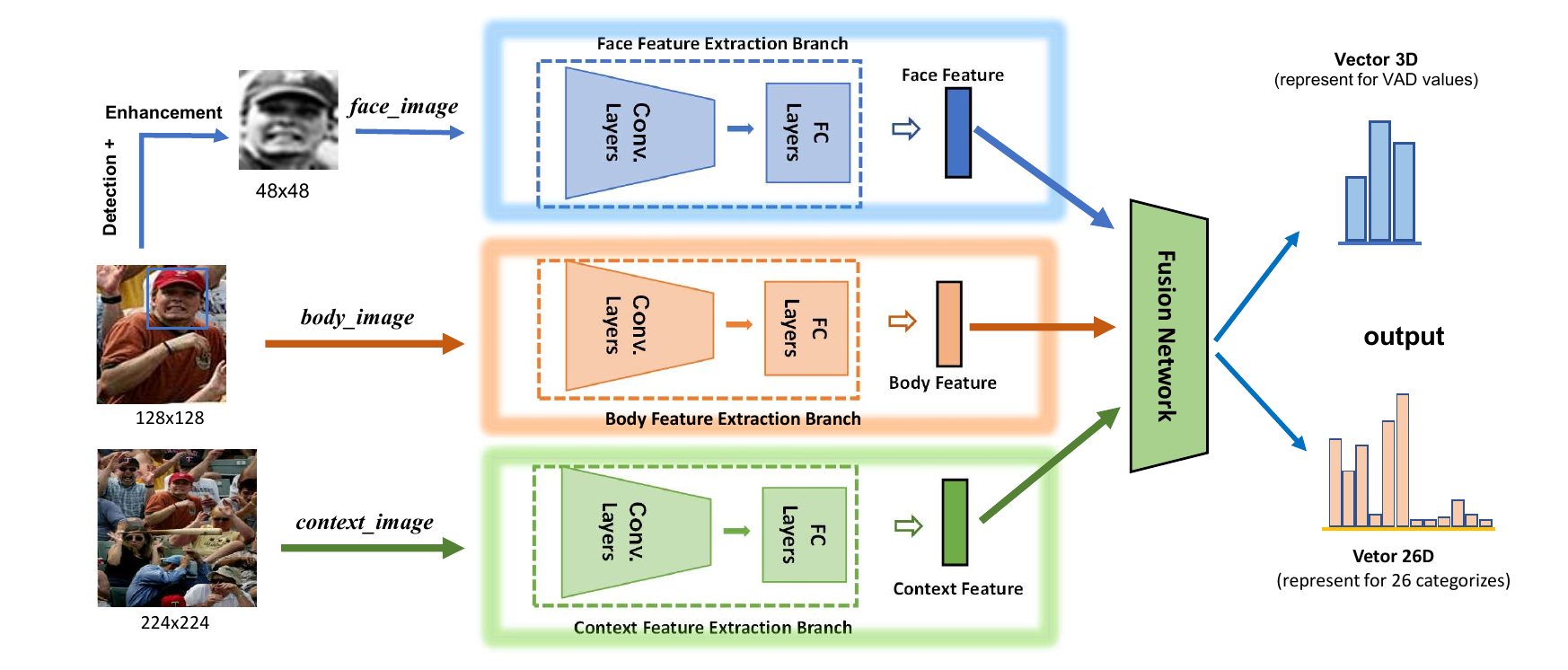}
    \caption{Architecture of our proposed Multi-Branch Network (MBN) for image emotion prediction.}
    \label{fig:proposeModel}
\end{figure*}

In recent methods for predicting the emotion of people in images, one of the most common approaches is combining facial expression and scene context information. The two most effective approaches are the EmotiCon algorithm ~\cite{mittal2020emoticon} and the fusion model ~\cite{8713881}. The fusion model consists of two feature extraction modules and a fusion network for jointly estimating the discrete categories and continuous dimensions. EmotiCon, which is based on Frege’s Principle, uses three interpretations of context to perform perceived emotion recognition. They use multiple modalities of faces and gaits, background visual information, and sociodynamic inter-agent interactions to infer the perceived emotion ~\cite{mittal2020emoticon}. Taking inspiration from the EmotiCon algorithm ~\cite{mittal2020emoticon} and the fusion model ~\cite{8713881}, we propose an multi-branch deep network for image emotion prediction.

\subsection{Discrete and Continuous Emotions}
Emotions are an important part of human life. They can help us better understand ourselves and those around us, as well as help us make informed decisions. Based on the existing literature in the fields of psychology and neuroscience, one’s emotional state can be described following two distinct representational approaches, namely the categorical and the dimensional approaches.

The categorical approach is based on research on basic emotions, pioneered by Darwin, interpreted by Tomkins, and supported by the findings of Ekman~\cite{ekman1987universals}. There are a small number of basic emotions, include happiness, sadness, fear, anger, surprise, and disgust. Within basic emotions, the term discrete emotion means that this emotion represents its own category. For example, basic theorists view fear, anger, and disgust as three separate discrete emotions. Rage, annoyance, and anger would be feelings that could be categorized within the discrete emotion of anger. 

But the 6 basic emotions cannot fully describe the full range of human emotions, so some new theories have divide into more types of emotions. A new study identifies 27 categories of emotion and shows how they blend together in our everyday experience~\cite{cowen2017self}. This findings will help other scientists and engineers more precisely capture the emotional states that underlie moods, brain activity, and expressive signals, leading to improved psychiatric treatments, an understanding of the brain basis of emotion, and technology responsive to our emotional needs~\cite{horikawa2020neural}. 

According to the dimensional approach, affective states are not independent of one another; rather, they are related to one another systematically. In this approach, the majority of affect variability is covered by three dimensions: valence, arousal, and potency (dominance) ~\cite{davitz1964expression} (see Fig.~\ref{fig:EmotionInVAD}). The valence dimension refers to how positive or negative the emotion is, and ranges from unpleasant feelings to pleasant feelings of happiness. The arousal dimension refers to how excited or apathetic the emotion is and it ranges from sleepiness or boredom to frantic excitement. The power dimension refers to the degree of power or sense of control over the emotion. The field concerned with the psychological study of emotion has recently witnessed much debate regarding the structure of emotion.

In particular, scientists have argued over whether emotions are best described along dimensions of valence and arousal or as discrete entities ~\cite{harmon2017importance}. However, we note that each approach has its own advantages and disadvantages. Therefore, we predict the emotion of an image based on both discrete emotion and VAD values.

\section{Propose Method}
\label{sec:III}

\subsection{Overview}

\begin{figure}[t!]
    \centering
    \includegraphics[width=\columnwidth]{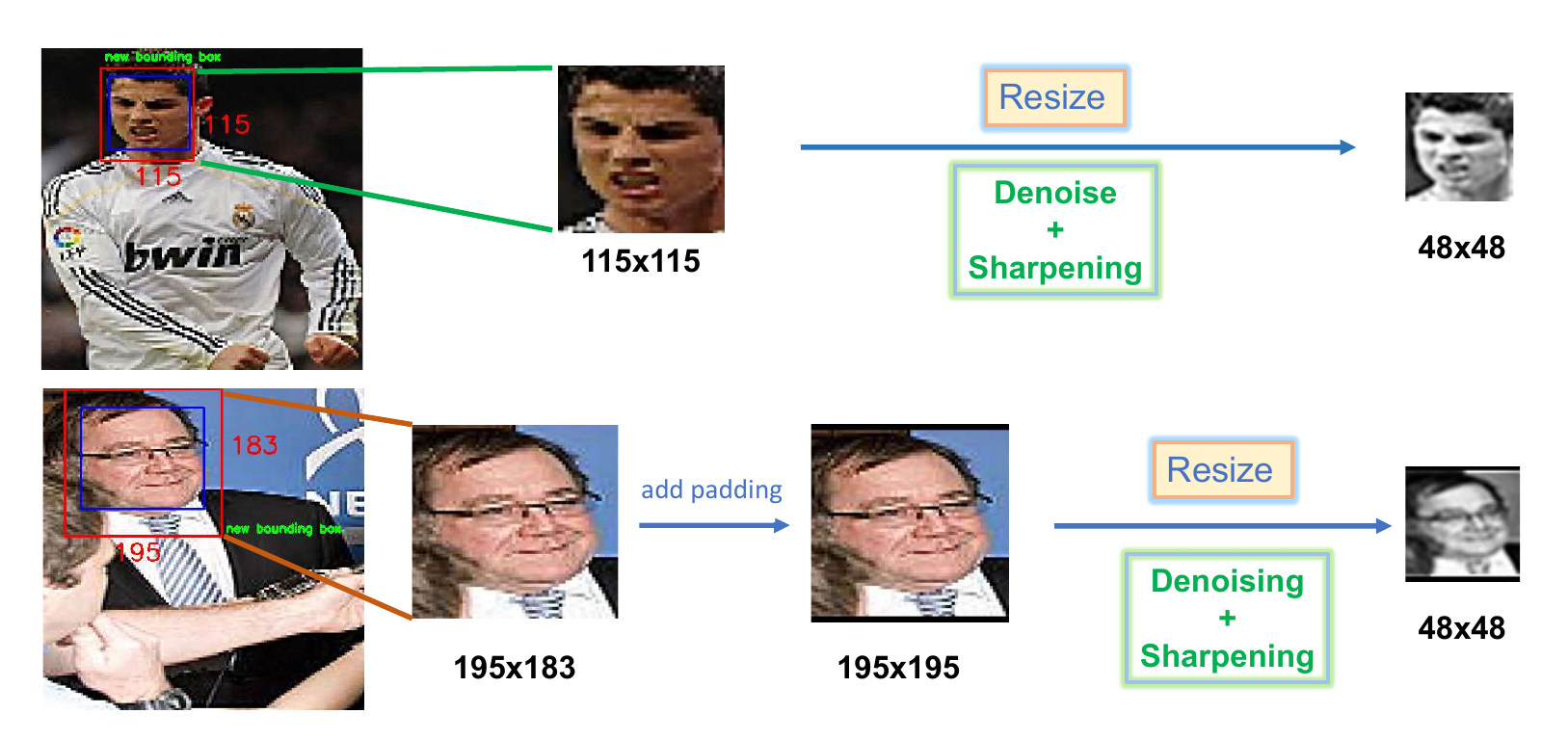}
    \caption{Face extraction and enhancement from \textit{body image}.}
    \vspace{-4mm}
    \label{fig:Process_Face_Image}
\end{figure}

Figure \ref{fig:proposeModel} shows the architecture of our proposed Multi-Branch Network (MBN). The network is divided into two main parts. The first part extracts features from the body, the face, and context of the image, referred to as \textit{body image}, \textit{face image}, and \textit{context image}. It consists of three branches to exploit emotions from subjects and background. We remark that the face region is extracted from the body image. The second part is a fusion network combining the features extracted from the three branches to predict the discrete emotions and VAD values of each person in the image.

% It consists of two branches corresponding to two pre-trained models that can predict effective information from images of subjects and background images, respectively. 

% The second part is used to extract features from the face that are detected from the body, referred to as , corresponding to a model that can recognize emotion from the face. 

\subsection{Feature Extraction Branches}

Our proposed network consists of three feature extraction branches, utilizing different deep learning models trained on suitable datasets to efficiently make predictions on human and scene images. 

% used to extract features from the body, context, and face of a person within an image. The Body Feature Extraction Branch and the Context Feature Extraction Branch are both deep learning models trained on suitable datasets to efficiently make predictions on human and scene images. The Face Feature Extraction Branch belongs to a CNN model trained on the FER-2013 dataset, which is used for facial emotion recognition. We removed the classifier layer of each model and froze their weights, and then used them to extract features from the \textit{body image}, \textit{context image} and \textit{face image}.

\textbf{Body Feature Extraction Branch. } This branch is designed to extract features from \textit{body image}. It takes an image that contains only the person's body, without the surrounding scene, and its objective is to extract critical information from the input. To achieve this goal, the backbone of the Body Feature Extraction Branch must work efficiently with images that contain only the person's body. Therefore, we use some deep learning backbones such as ResNet-18, ResNet-50~\cite{he2016deep}, and SwinT\cite{liu2021swin}, which come with ImageNet\cite{deng2009imagenet} pre-trained weights since ImageNet is a large-scale dataset of human images. Additionally, we also use some self-retrained models. We retrained the ResNet-50 and SwinT models using the EMOTIC dataset. We utilize these backbones to build a distinct model from scratch for predicting the gender and age range of individuals in the image. These models were trained for 15 epochs, employing the \textit{body image} of the EMOTIC dataset as both the training and testing datasets. The results showed that both models achieved an mAP score above 0.33, indicating satisfactory performance. %We saved these models to be used for feature extraction from \textit{body images} in the future.
%Both models achieved an mAP above 0.33 in recognizing the gender and age range of the person in the image, which is good enough to demonstrate the effectiveness of the model. 
We believe that 
% models performing well on the body-only image dataset can extract useful information from \textit{body image}, and 
retraining these models on the EMOTIC dataset can improve their effectiveness.

\textbf{Context Feature Extraction Branch. } This branch is utilized to extract features from the scene context of an image. Similarly, we use backbones that perform well in scene images for this branch. To achieve the goal, we use the ResNet-18 and ResNet-50 models with weights pre-trained on the Places365 dataset~\cite{zhou2017places}, which consists of more than 1.8 million images from 365 scene classes.

\textbf{Face Feature Extraction Branch. } The face branch is one of the most essential parts of our proposed network and it represents our additional improvement over the previous approach~\cite{8713881}. As we mentioned, the key to predicting emotion in a human image is facial expression. Therefore, we utilize different deep learning models pre-trained on FER-2013 dataset~\cite{FER2013}, a facial expression recognition dataset consisting of 30,000 facial images with 7 different expressions: angry, disgust, fear, happy, sad, surprise, and neutral. The FER-2013 dataset is also a popular benchmark for facial expression recognition algorithms. Particularly, we employ two pre-trained models on the FER-2013 dataset by Shangeth~\cite{shangeth2022facial} and Balmukund~\cite{balmukundfacial}, denoted by S-FER and B-FER, respectively.

\textbf{Face Extraction and Enhancement. } We employ a face detection~\cite{facedetection} method to locate the bounding box of the face within the \textit{body image} and then crop the image accordingly. Subsequently, we resize the cropped images to 48x48. %Finally, we enhance the quality of the facial images through some simple processing steps, including denoising and sharpening.

% Because the EMOTIC dataset does not include facial images of person whose emotions are to be predicted, when proposing to incorporate additional information extracted from faces, we had to create a facial image set. First, we used a face detection model to locate the facial regions and cropped the facial images with a 1:1 aspect ratio to preserve image quality during resizing. Subsequently, we resized the cropped images to 48x48 to match the FER model trained on the FER-2013 dataset. Finally, we enhanced the quality of the facial images through some simple processing steps, including denoising and sharpening, before conducting the training process.

% Due to the unavailability of direct facial images in the EMOTIC dataset, we extract facial images from the \textit{body image} for input into the Face Feature Extraction Branch. To accomplish this, we employ a face detection model to locate the bounding box of the face within the \textit{body image} and then crop the image accordingly. As the Face Feature Extraction Branch's backbones are designed for the FER-2013 dataset, we resize the input images to 48x48 after cropping.

As \textit{body image} was not originally of high quality, resizing \textit{face image} to different height and width ratios after cropping may significantly reduce the quality of the facial image. To ensure optimal results from the Face Feature Extraction Branch, we first extend the bounding box to ensure equal width and height before cropping. In cases where the extension is not possible (e.g., faces near the image edges), we add padding after cropping to achieve equal width and height. These small preprocessing steps ensure that the \textit{face image} are preserved with improved quality.

Nevertheless, since the extracted facial images from the \textit{body image} still have suboptimal quality, it necessary to enhance their quality before inputting them into the Face Feature Extraction Branch. So, after cropping and resizing the face images to 48x48, we further improve their quality by performing denoising and sharpening(see Fig. \ref{fig:Process_Face_Image}). We use the \textit{filter2D} function of \textit{OpenCV} to make the \textit{face image} appear sharper and maintain its quality.

\subsection{Fusion Network}

Architecture of the fusion network is adapted from the work of Ronak et al.~\cite{8713881}, which is combined of two fully-connected layers. The first fully-connected layer is to reduce the dimension of the input, which is taken from the last layer of the predecessor models to 256, and then the second fully-connected layer is used to predict the discrete emotions and VAD value of the emotions. 

To connect the fusion network with the three predecessors, we cut off the last layer of all the element models, then rebuild the first three by creating three sequential networks of all the layers of each original model, except for the last being cut off. Then we pass the number of input features from the last layer of three models into a linear layer to construct the first fully-connected layer of the third part. The output of this layer is a vector with size of 256. The second layer is a linear layer, which takes in a vector with size of 256 and outputs two vectors, first has 26 dimensions denoting 26 kinds of discrete emotions, and the second has 3 dimensions denoting VAD value.

% \subsection{Detect face and enhance face image}

\subsection{Loss function}
\label{subsec:loss_function}
\subsubsection{Criterion for Discrete Categories}
The problem of predicting discrete emotions for people in images is a multilabel prediction task, because each person is labeled with multiple different emotions from among the N kinds of discrete emotions. Additionally, the issue of data imbalance (see Fig. \ref{fig:Distribution}) poses difficulties in training a model to produce the most accurate predictions.

Following instructions of the EMOTIC dataset~\cite{8713881}, we employ a weighted Euclidean loss during the training process. This loss function is deemed more appropriate for the task than Kullback-Leibler divergence or a multi-class multi-classification hinge loss~\cite{8713881}. Concretely, for each prediction $\hat{y} = [\hat{y}_1, \hat{y}_2,..., \hat{y}_{N}]$, the weighted Euclidean loss is defined as follows:
\begin{equation}
    L^{disc}(\hat{y}) = \sum_{i=1}^{N} w_{i}(\hat{y}_{i} - y_{i})^2,
\end{equation}
where N is the number of categories, $\hat{y}_{i}$ is the prediction for the $i^{th}$ class and $y_{i}$ is the true label of this class. The parameter $w_{i}$ is the weight assigned to each class.  $w_{i} = \frac{1}{ln(c+p_{i})}$, where $p_{i}$ is the propability of the class $i^{th}$ and c is the parameter to control the range of valid values for $w_{i}$.
\subsubsection{Criterion for Continuous Dimensions} The task of predicting values on a continuous scale is formally treated as a regression problem. Consequently, the adoption of L2 Loss as the preferred loss function is deemed appropriate for conducting this particular task. L2 Loss (Euclidean Loss or Mean Squared Error) is a common loss function in machine learning used for regression tasks. It calculates the squared difference between predicted and actual values, aiming to minimize overall error and create a smoother model output. By penalizing large errors more, it helps achieve greater accuracy and precision in regression models. The optimization process aims to find the model's parameters that result in the minimum L2 Loss, which leads to a more accurate and precise model for regression tasks. Concretely, for each prediction $\hat{y} = [\hat{y}_1, \hat{y}_2,..., \hat{y}_{N}]$, the L2 loss is defined as follows:
\begin{equation}
    L^{cont}(\hat{y}) = \sum_{i=1}^{N} v_{i}(\hat{y}_{i} - y_{i})^2,
\end{equation}
where N is the number of dimensions, $\hat{y}_{i}$ is the prediction for the $i^{th}$ dimension value and $y_{i}$ is the true label of this dimension. The parameter $v_{i} \in \{0,1\}$ is a binary weight to represent the error margin. $v_{i} = 0$ if  $(\hat{y}_{i} - y_{i}) < \theta$ else $v_{i} = 1$. If the prediction of each dimension is too close to the ground truth, i.e., the error is less than $\theta$, then the loss for this dimension is effectively crossed out or set to zero.

\section{EXPERIMENTS}
\label{sec:IV}

\subsection{Implementation Details}
We used Google Colab as the platform to conduct experiments, taking advantage of the free resources it provided. Specifically, we had access to 78GB of disk space, 12GB of RAM, and 12GB of T4 GPU RAM. Our deep learning model was implemented with the PyTorch library. We conducted the experiments using a batch size of 52 for the discrete emotion prediction tasks and a batch size of 26 for the continuous emotion prediction tasks. Our model trained for 15 epochs with an initial learning rate of 0.001. We conducted tests with various initial learning rates and determined that 0.001 is the most suitable one. The learning rate was decayed by 10\% every 7 epochs to prevent overfitting. The optimization algorithm used was stochastic gradient descent (SGD) with momentum of 0.9.

\subsection{Dataset}
We evaluate methods on the EMOTIC (i.e., EMOTions In Context) dataset~\cite{8713881}, a database of images of people in real environments, annotated with their apparent emotions. The images are annotated with an extended list of 26 emotion categories combined with the three common continuous dimensions Valence, Arousal and Dominance. It consists of approximately 23,500 images collected from websites, social media, and other public datasets. Each image contains one or many people, and each person is labeled with Gender (Male or Female), Age (adult, kid, or teenager), a bounding box (top-left, bottom-right) of full body and VAD (valence-arousal-dominance) values ranging from 0 to 10 and labeled to some of 26 discrete emotion categories. The 26 discrete emotion categories include: \textit{Peace, Affection, Esteem, Anticipation, Engagement, Confidence, Happiness, Pleasure, Excitement, Surprise, Sympathy, Doubt/Confusion, Disconnection, Fatigue, Embarrassment, Yearning, Disapproval, Aversion, Annoyance, Anger, Sensitivity, Sadness, Disquietment, Fear, Pain, Suffering.}

\begin{figure}[t!]
  \centering
  \begin{subfigure}[b]{\columnwidth}
    \includegraphics[width=\columnwidth]{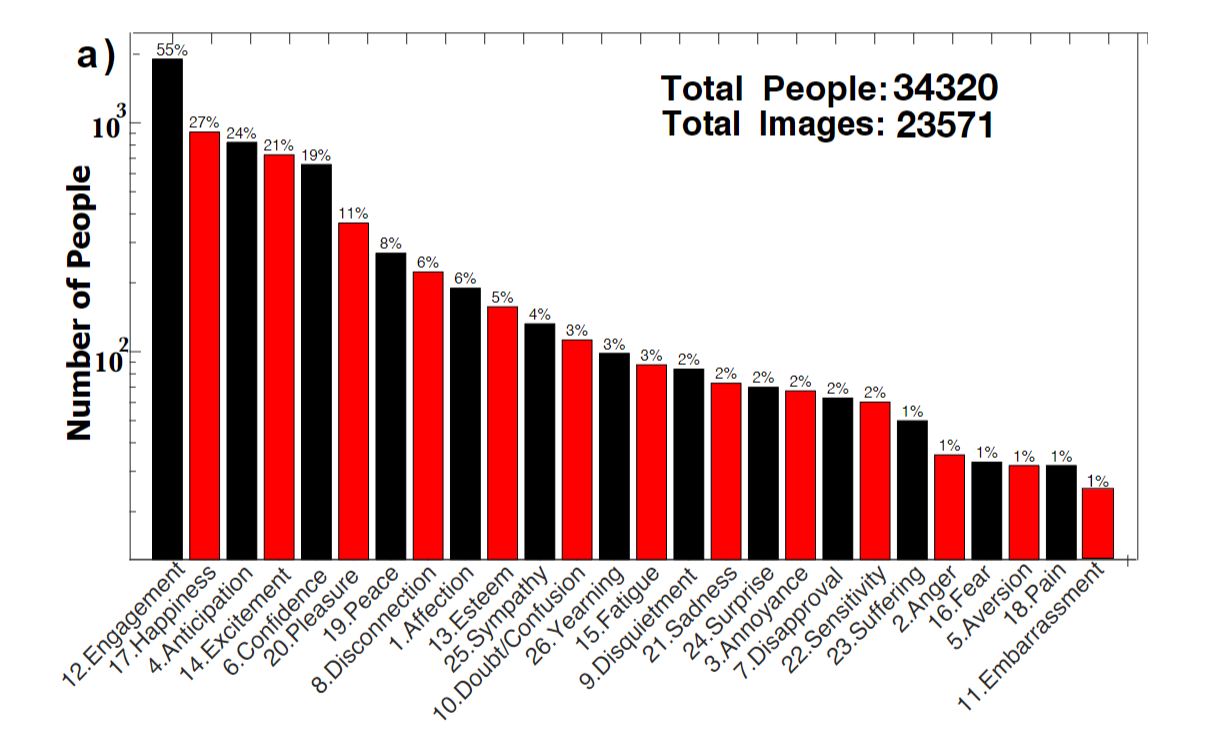}
  \end{subfigure}\\
  \begin{subfigure}[b]{\columnwidth}
    \includegraphics[width=\columnwidth]{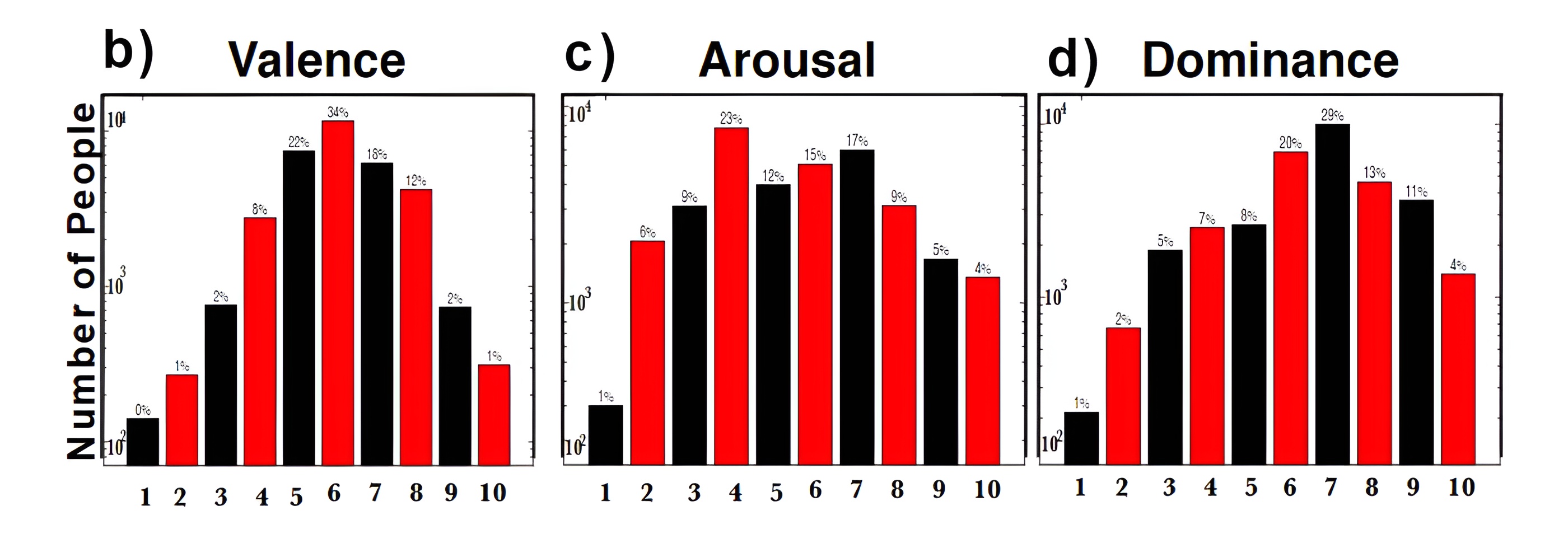}
  \end{subfigure}
  \caption{ Dataset Statistics. (a) Number of people annotated for
each emotion category; (b), (c) and (d) Number of people annotated for every value of the three continuous dimensions : Valence, Arousal and Dominance ~\cite{8713881}.}
    \label{fig:Distribution}
\end{figure}

In this paper, we conduct the experiment on both the 26 discrete emotion categories and the VAD continuous emotion value. Figure \ref{fig:Distribution} shows the distribution of the number of people annotated across 26 emotion categories and three continuous dimensions. The discrete distribution is uneven, with Engagement being the most common (about 55\%) and Embarrassment being the least common (about 1\%). While the VAD value is annotated quite evenly. 

With a large number of people with various emotional states in the dataset, deep learning models can learn to recognize more complex patterns and relationships in the data, which can improve the accuracy and predictive power. However, EMOTIC's considerable imbalance (see Fig. \ref{fig:Distribution}) adversely impacts the models' learning performance for emotions with limited samples. As a result, the models excel at recognizing emotions that have a higher number of samples.

\subsection{Evaluate Metric}
\subsubsection{Discrete Prediction Evaluation Metric}
The metric used to evaluate the prediction of the discrete results is mAP (mean Average Precision). mAP is used to evaluate the model's performance in multi-class classification tasks. The higher mAP value, the better the model achieves in object recognition.

The AP for an object class is calculated as follows:
\begin{equation}
AP = \sum_n (R_n - R_{n-1}) \cdot P_n,
\end{equation}
where where $P_n$ and $R_n$ are the precision and recall at the $n^{th}$ threshold. For inference, we chose the threshold for each class at which the precision and recall are equal.

To calculate mAP, we calculate AP for each emotion category and take the average of those AP values. The mAP is calculated as follows:
\begin{equation}
mAP = \frac{1}{N} \sum_{i=1}^N AP_i,
\end{equation}
where $N$ is the number of categories. $AP_i$ is the AP value of this discrete emotion prediction.

\subsubsection{Continuous Prediction Evaluation Metric}
For the continuous dimension emotion VAD, we evaluate the result by the mean absolute error (MAE), which is the mean absolute error regression loss, which one lower is the better. The MAE value for each dimension is calculated as below:
\begin{equation}
MAE = \frac{1}{N} \sum_{i=1}^N (|\hat{y_i} - y_i|),
\end{equation}
where $\hat{y_i}$ denotes the prediction value for a dimension of the $i^{th}$ image and $y_i$ is the true continuous value for that dimension of the image number $i$. The MAE of N dimensions is then calculated as the mean of N values of MEA of each dimension.

\subsection{Experiment Results}

% % Please add the following required packages to your document preamble:
% % \usepackage{multirow}
% % \usepackage{graphicx}
\begin{table}[t!]
\caption{Discrete and continuous results using only body or context or facial feature.}
\resizebox{\columnwidth}{!}{%
\begin{tabular}{|ccc|c|c|}
\hline
\multicolumn{3}{|c|}{\textbf{BACKBONE}} &
  \multirow{2}{*}{\textbf{mAP}} &
  \multirow{2}{*}{\textbf{MAE}} \\ \cline{1-3}
\multicolumn{1}{|c|}{\textit{\textbf{Body Branch}}} &
  \multicolumn{1}{c|}{\textit{\textbf{Context Branch}}} &
  \textit{\textbf{Face Branch}} &
   &
   \\ \hline
\multicolumn{1}{|c|}{\begin{tabular}[c]{@{}c@{}}Resnet-18\\ (ImageNet weight)\end{tabular}} &
  \multicolumn{1}{c|}{-} &
  - &
  0.2376 &
  0.9757 \\ \hline
\multicolumn{1}{|c|}{\begin{tabular}[c]{@{}c@{}}Resnet-50\\ (ImageNet weight)\end{tabular}} &
  \multicolumn{1}{c|}{-} &
  - &
  0.2465 &
  0.9606 \\ \hline
\multicolumn{1}{|c|}{\begin{tabular}[c]{@{}c@{}}Resnet-50\\ (Fine-tunning on EMOTIC)\end{tabular}} &
  \multicolumn{1}{c|}{-} &
  - &
  0.2445 &
  0.9590 \\ \hline
\multicolumn{1}{|c|}{\begin{tabular}[c]{@{}c@{}}SwinT\\ (ImageNet weight)\end{tabular}} &
  \multicolumn{1}{c|}{-} &
  - &
  \textbf{0.2671} &
  0.9403 \\ \hline
\multicolumn{1}{|c|}{\begin{tabular}[c]{@{}c@{}}SwinT\\ (Fine-tunning on EMOTIC)\end{tabular}} &
  \multicolumn{1}{c|}{-} &
  - &
  0.2666 &
  \textbf{0.9356} \\ \hline
\multicolumn{1}{|c|}{-} &
  \multicolumn{1}{c|}{\begin{tabular}[c]{@{}c@{}}Resnet-18\\ (Places365 weight)\end{tabular}} &
  - &
  0.2538 &
  \multicolumn{1}{l|}{0.9673} \\ \hline
\multicolumn{1}{|c|}{-} &
  \multicolumn{1}{c|}{\begin{tabular}[c]{@{}c@{}}Resnet-50\\ (Places365 weight)\end{tabular}} &
  - &
  \textbf{0.2598} &
  \multicolumn{1}{l|}{0.9616} \\ \hline
\multicolumn{1}{|c|}{-} &
  \multicolumn{1}{c|}{-} &
  S-FER &
  0.1840 &
  1.0385 \\ \hline
\multicolumn{1}{|c|}{-} &
  \multicolumn{1}{c|}{-} &
  B-FER &
  \textbf{0.2061} &
  \textbf{1.0259} \\ \hline
\end{tabular}%
}
\label{table:one-branch}
\end{table}

%%%%%%%%%%%%%%%%%%%%%%%%%%%%%%%%%%%%

% \usepackage{multirow}
% \usepackage{graphicx}
\begin{table}[t!]
\caption{Discrete and continuous results using a combination of body, context, and facial features.}
\resizebox{\columnwidth}{!}{%
\begin{tabular}{|ccc|c|c|}
\hline
\multicolumn{3}{|c|}{\textbf{BACKBONE}} &
  \multirow{2}{*}{\textbf{mAP}} &
  \multirow{2}{*}{\textbf{MAE}} \\ \cline{1-3}
\multicolumn{1}{|c|}{\textit{\textbf{Body Branch}}} &
  \multicolumn{1}{c|}{\textit{\textbf{Context Branch}}} &
  \textit{\textbf{Face Branch}} &
   &
   \\ \hline
\multicolumn{1}{|c|}{\begin{tabular}[c]{@{}c@{}}Resnet-18\\ (ImageNet weight)\end{tabular}} &
  \multicolumn{1}{c|}{\begin{tabular}[c]{@{}c@{}}Resnet-18\\ (Places365 weight)\end{tabular}} &
  - &
  0.2585 &
  0.9615 \\ \hline
\multicolumn{1}{|c|}{\begin{tabular}[c]{@{}c@{}}Resnet-50\\ (ImageNet weight)\end{tabular}} &
  \multicolumn{1}{c|}{\multirow{4}{*}{\begin{tabular}[c]{@{}c@{}}Resnet-50\\ (Places365 weight)\end{tabular}}} &
  - &
  0.2645 &
  0.9525 \\ \cline{1-1} \cline{3-5} 
\multicolumn{1}{|c|}{\begin{tabular}[c]{@{}c@{}}Resnet-50\\ (Fine-tunning on EMOTIC)\end{tabular}} &
  \multicolumn{1}{c|}{} &
  - &
  0.2640 &
  0.9483 \\ \cline{1-1} \cline{3-5} 
\multicolumn{1}{|c|}{\begin{tabular}[c]{@{}c@{}}SwinT\\ (ImageNet weight)\end{tabular}} &
  \multicolumn{1}{c|}{} &
  - &
  \textbf{0.2774} &
  \textbf{0.9411} \\ \cline{1-1} \cline{3-5} 
\multicolumn{1}{|c|}{\begin{tabular}[c]{@{}c@{}}SwinT\\ (Fine-tunning on EMOTIC)\end{tabular}} &
  \multicolumn{1}{c|}{} &
  - &
  0.2756 &
  0.9422 \\ \hline
\multicolumn{1}{|c|}{\multirow{2}{*}{-}} &
  \multicolumn{1}{c|}{\multirow{2}{*}{\begin{tabular}[c]{@{}c@{}}Resnet-18\\ (Places365 weight)\end{tabular}}} &
  S-FER &
  0.2590 &
  \textbf{0.9419} \\ \cline{3-5} 
\multicolumn{1}{|c|}{} &
  \multicolumn{1}{c|}{} &
  B-FER &
  0.2615 &
  0.9461 \\ \hline
\multicolumn{1}{|c|}{\multirow{2}{*}{-}} &
  \multicolumn{1}{c|}{\multirow{2}{*}{\begin{tabular}[c]{@{}c@{}}Resnet-50\\ (Places365 weight)\end{tabular}}} &
  S-FER &
  0.2659 &
  0.9435 \\ \cline{3-5} 
\multicolumn{1}{|c|}{} &
  \multicolumn{1}{c|}{} &
  B-FER &
  \textbf{0.2674} &
  0.9471 \\ \hline
\multicolumn{1}{|c|}{\begin{tabular}[c]{@{}c@{}}Resnet-18\\ (ImageNet weight)\end{tabular}} &
  \multicolumn{1}{c|}{-} &
  \multirow{2}{*}{S-FER} &
  0.2467 &
  0.9545 \\ \cline{1-2} \cline{4-5} 
\multicolumn{1}{|c|}{\begin{tabular}[c]{@{}c@{}}Resnet-50\\ (ImageNet weight)\end{tabular}} &
  \multicolumn{1}{c|}{-} &
   &
  \textbf{0.2554} &
  \textbf{0.9455} \\ \hline
\end{tabular}%
}
\label{table:two-branch}
\end{table}
%%%%%%%%%%%%%%%%%%%%%%%%%%%%%%%%

% \usepackage{multirow}
% \usepackage{graphicx}
\begin{table}[t!]
\caption{Discrete and continuous results using body, context, and facial features.}
\resizebox{\columnwidth}{!}{%
\begin{tabular}{|ccc|c|c|}
\hline
\multicolumn{3}{|c|}{\textbf{BACKBONE}} &
  \multirow{2}{*}{\textbf{mAP}} &
  \multirow{2}{*}{\textbf{MAE}} \\ \cline{1-3}
\multicolumn{1}{|c|}{\textit{\textbf{Body Branch}}} &
  \multicolumn{1}{c|}{\textit{\textbf{Context Branch}}} &
  \textit{\textbf{Face Branch}} &
   &
   \\ \hline
\multicolumn{1}{|c|}{\multirow{2}{*}{\begin{tabular}[c]{@{}c@{}}Resnet-18\\ (ImageNet weight)\end{tabular}}} &
  \multicolumn{1}{c|}{\multirow{2}{*}{\begin{tabular}[c]{@{}c@{}}Resnet-18\\ (Places365 weight)\end{tabular}}} &
  S-FER &
  0.2638 &
  0.9491 \\ \cline{3-5} 
\multicolumn{1}{|c|}{} &
  \multicolumn{1}{c|}{} &
  B-FER &
  0.2591 &
  0.9479 \\ \hline
\multicolumn{1}{|c|}{\multirow{2}{*}{\begin{tabular}[c]{@{}c@{}}Resnet50\\ (ImageNet weight)\end{tabular}}} &
  \multicolumn{1}{c|}{\multirow{2}{*}{\begin{tabular}[c]{@{}c@{}}Resnet-50\\ (Places365 weight)\end{tabular}}} &
  S-FER &
  0.2689 &
  0.9349 \\ \cline{3-5} 
\multicolumn{1}{|c|}{} &
  \multicolumn{1}{c|}{} &
  B-FER &
  0.2726 &
  0.9465 \\ \hline
\multicolumn{1}{|c|}{\multirow{2}{*}{\begin{tabular}[c]{@{}c@{}}Resnet-50\\ (Fine-tuning on EMOTIC)\end{tabular}}} &
  \multicolumn{1}{c|}{\multirow{2}{*}{\begin{tabular}[c]{@{}c@{}}Resnet-50\\ (Places365 weight)\end{tabular}}} &
  S-FER &
  0.2731 &
  0.9337 \\ \cline{3-5} 
\multicolumn{1}{|c|}{} &
  \multicolumn{1}{c|}{} &
  B-FER &
  0.2721 &
  0.9478 \\ \hline
\multicolumn{1}{|c|}{\multirow{2}{*}{\begin{tabular}[c]{@{}c@{}}SwinT\\ (ImageNet weight)\end{tabular}}} &
  \multicolumn{1}{c|}{\multirow{2}{*}{\begin{tabular}[c]{@{}c@{}}Resnet-50\\ (Places365 weight)\end{tabular}}} &
  S-FER &
  0.2821 &
  0.9278 \\ \cline{3-5} 
\multicolumn{1}{|c|}{} &
  \multicolumn{1}{c|}{} &
  B-FER &
  0.2828 &
  0.9296 \\ \hline
\multicolumn{1}{|c|}{\multirow{2}{*}{\begin{tabular}[c]{@{}c@{}}SwinT\\ (Fine-tunning on EMOTIC)\end{tabular}}} &
  \multicolumn{1}{c|}{\multirow{2}{*}{\begin{tabular}[c]{@{}c@{}}Resnet-50\\ (Places365 weight)\end{tabular}}} &
  S-FER &
  \textbf{0.2837} &
  0.9323 \\ \cline{3-5} 
\multicolumn{1}{|c|}{} &
  \multicolumn{1}{c|}{} &
  B-FER &
  0.2829 &
  \textbf{0.9256} \\ \hline
\end{tabular}%
}
\label{table:three-branch}

\end{table}

 \begin{figure*}[t!]
    \centering
\includegraphics[width = \textwidth]{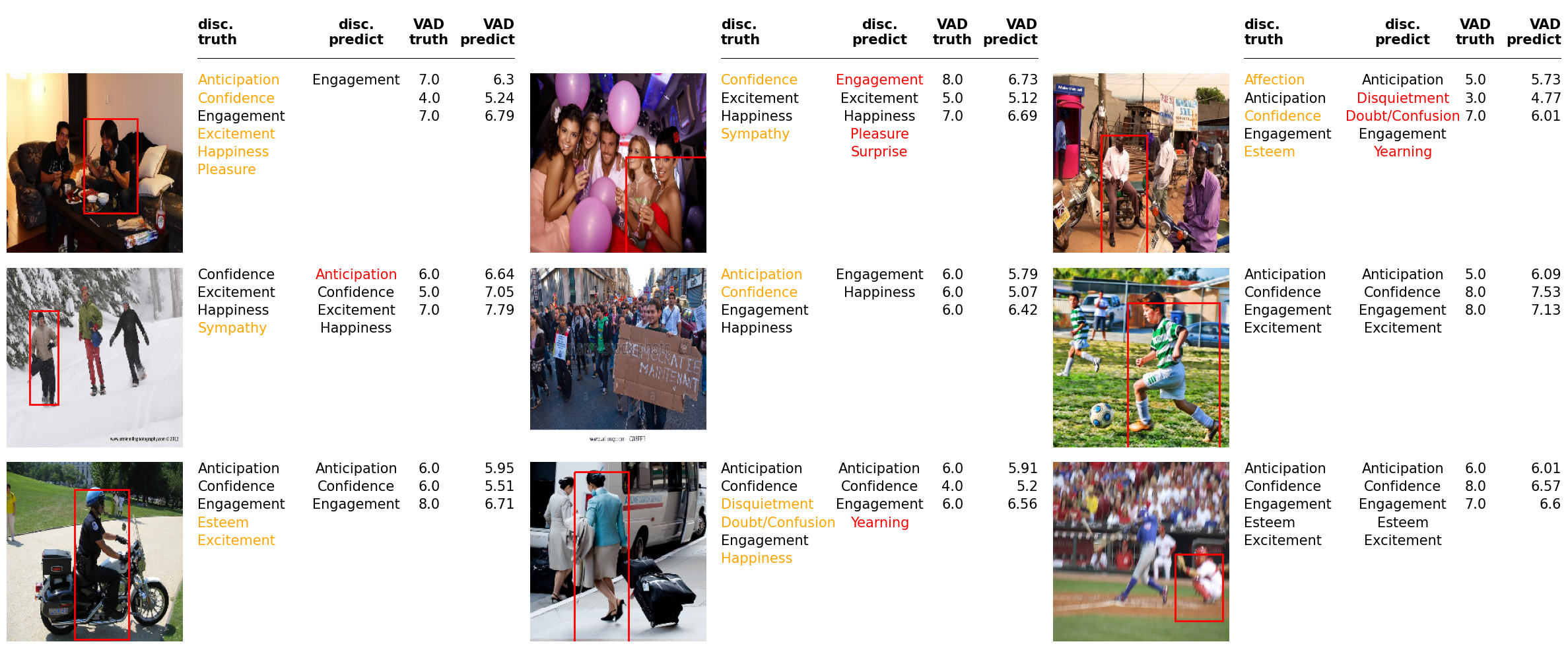}
    \caption{Visualization of our results. Images are chosen randomly with corresponding ground-truth and predicted discrete and continuous emotion.}
    \label{fig:enter-label}
\end{figure*}

To demonstrate the effectiveness of our proposed method, we conducted experiments using various backbones. Concretely, Body Feature Extraction Branch used Resnet-18, Resnet-50, and SwinT; Context Feature Extraction Branch used Resnet-18 and Resnet-50; and Face Feature Extraction Branch used S-FER and B-FER. For each experiment, we performed predictions for both discrete and continuous emotions. First, experiments were conducted using only one type of image information to investigate the effectiveness of each backbone and the amount of information contained in each type of image, as shown in Table \ref{table:one-branch}. Next, we used combined information from two out of three types of images to demonstrate the effectiveness of combining more information, as well as to provide a baseline for comparison with our proposed method, and verified the effectiveness of the previously method~\cite{8713881}, as shown in Table \ref{table:two-branch}. Finally, we tested our proposed model on all three types of information, as shown in Table \ref{table:three-branch}.

\subsubsection{Discretes Emotion Prediction}
Table \ref{table:one-branch} shows that using information from \textit{body image} or \textit{context image} results in superior performance compared to using information from \textit{face image}, due to the lower quality of facial images used in our experiments. The Body Feature Extraction Branch with the SwinT backbone achieved the highest mAP score of 0.2671. Despite this, the Context Feature Extraction Branch still performs well with an mAP score above 0.25, offering more consistent results without requiring a large model such as SwinT.

The results presented in Table \ref{table:two-branch} provide evidence that combining multiple types of information yields better performance than using individual information alone, resulting in an average increase of 11.2\% in mAP score. Additionally, stronger backbones were found to produce better results for each experiment. These experiments also demonstrate that all three types of images contain significant information for predicting emotions, as combining any two types of images consistently improved performance. When using two types of information, the method presented in ~\cite{8713881}, which combines information extracted from \textit{body image} and \textit{context image}, yields the best results.

Combining information from all three types of images significantly improved prediction results, with an mAP score above 0.28, using powerful models including SwinT, Resnet-50, and FER models. Beside, combining different backbones also proves effective, with an mAP score above 0.26. Compared to using only two types of information, the results are improved by 4.1\%. 

\subsubsection{Continous Emotion Prediction}
The results for continuous emotion prediction followed the same patterns as discrete emotion prediction. Specifically, MAE decreased by 3.5\% when using two types of information compared to using only one type, and decreased by 1.1\% when using three types of information compared to using two types. Moreover, our proposed model achieving the remarkable result with the lowest MAE of 0.9256 by combining SwinT, Resnet-50, and B-FER. These findings demonstrate the importance of combining multiple types of contextual information and powerful models for improving emotion prediction.
\subsubsection{Overall}
 Our proposed method outperformed the approach presented in ~\cite{8713881}, which utilizes a combination of information from \textit{body image} and \textit{context image}, in both discrete and continuous emotion prediction. Specifically, the mAP score increased by 2.3\% and the MAE score decreased by 1.2\%. Beside, our study demonstrated that all three types of images (e.g., \textit{body image}, \textit{context image} and \textit{face image}) contain important information for emotion prediction. Additionally, utilizing appropriate and stronger backbones leads to more accurate results.

\section{Conclusion}
\label{sec:V}

In summary, this paper provides a Multiple -Branch Network that combines information from the body, face, and context to enhance emotion prediction from images. Our results on the EMOTIC dataset indicate that combining body, face, and context images improves emotion prediction accuracy than just by using one or two kinds of features. Our work also contributes by utilizing the cutting-edge CNN models with pre-train weights on available datasets and applying them to emotion prediction. We also provide further techniques to optimize CNN's accuracy for facial features extraction, such as denoising and sharpening.  Our proposed method achieved mAP scores above 0.26 and MAE scores below 0.95. In additional, utilizing powerful deep learning models like SwinT, ViT, etc., with our approach yields significant results. The Swin Transformer backbone performed best with a mAP of 0.2837 and a MAE of 0.9256, but requires more computational time due to its larger size.

Although we made efforts, this work has limitations. The inconsistent quality of the dataset, which is the unequal in emotion distribution, could cause the learning process into bias. Due to limited resources, we could only conduct experiments on a limited number of models and datasets, which may lead to incomplete or inaccurate evaluations. Moreover, the image processing and the feature extraction techniques used were still in their early stages. Each extraction model could be improved by training them to learn more in-depth information relevant to their features, such as the body gestures for the body branch or place attributes for the context branch, etc. Future research can address these issues for better results.

\begin{acks}
This research is supported by research funding from Faculty of Information Technology, University of Science, Vietnam National University - Ho Chi Minh City.\\
We would like to express our gratitude to Dr. Minh-Triet Tran for his valuable suggestions and support that greatly contributed to the completion of this research.
\end{acks}

\bibliographystyle{ACM-Reference-Format}
\bibliography{refs}

\end{document}